\documentclass{ecai}
\usepackage{times}
\usepackage{latexsym}
\usepackage{amssymb}
\usepackage[numbers, square, sectionbib]{natbib}
\usepackage{color}
\usepackage[ruled,linesnumbered,boxed]{algorithm2e}
\usepackage{mathtools}
\usepackage{amsthm}
\usepackage{soul}
\usepackage{bm}

\usepackage{graphicx}
\usepackage{mathtools}
\SetKwInOut{Parameter}{Parameters}
\usepackage{url}

\usepackage{mwe,graphicx,tabu,subcaption,float}
\usepackage{color}

\DeclarePairedDelimiterX{\infdivx}[2]{(}{)}{%
  #1\;\delimsize\|\;#2%
}

\SetKwComment{Comment}{$\triangleright$\ }{}

\begin{document}

\title{Feature Importance Estimation with Self-Attention Networks }

\author{Bla\v{z} \v{S}krlj$^1$ \and Sa\v{s}o D\v{z}eroski$^1$ \and Nada Lavra\v{c}$^1$ \and Matej Petkovi\'{c}\footnote{Jo\v{z}ef Stefan Institute and International Postgraduate School Jo\v{z}ef Stefan, Slovenija, matej.petkovic@ijs.si}}

\maketitle
\bibliographystyle{ecai}

\begin{abstract}
Black-box neural network models are widely used in industry and science, yet are hard to understand and interpret.
Recently, the attention mechanism was introduced, offering insights into the inner workings of neural language models.
This paper explores the use of attention-based neural networks mechanism for estimating feature importance, as means for explaining the models learned from propositional (tabular) data. Feature importance estimates, assessed by the proposed Self-Attention Network (SAN) architecture, are compared with the established ReliefF, Mutual Information and Random Forest-based estimates, which are widely used in practice for model interpretation. 
For the first time we conduct scale-free comparisons of feature importance estimates across algorithms on ten real and synthetic data sets to study the similarities and differences of the resulting feature importance estimates, showing that SANs identify similar high-ranked features as the other methods. 
We demonstrate that SANs identify feature interactions which in some cases yield better predictive performance than the baselines, suggesting that attention extends beyond interactions of just a few key features and detects larger feature subsets relevant for the considered learning task.
\end{abstract}

\section{Introduction}
\label{sec:intro}
Deep neural networks have been successfully applied to text, graph and image-based classification tasks, as well as to learning accurate classifiers from propositional (tabular) data \cite{deng2014deep,goodfellow2016deep,lecun2015deep}.
However, with increasing number of parameters, neural networks are becoming less human-understandable. The currently adopted paradigm to tackle this problem involves using \emph{post hoc} explanation tools, such as SHAP and LIME \cite{lundberg2017unified}. Such approaches operate by approximating the outputs of a given black-box model via some tractable scheme, such as efficient computation of Shapley values or local approximation via a linear model. Such methods do not take into account the inner structure of a given (deep) neural network, and treat it as a black box, only considering its inputs and outputs.

Recent advances in the area of language processing, however, offer the opportunity to link parts of a given neural network with the input space directly, via the \emph{attention mechanism}
\cite{vaswani2017attention}. Achieving super-human performance on many language understanding tasks, attention-based models are becoming widely adopted throughout scientific and industrial environments \cite{wang2019superglue}. Models, such as BERT \cite{devlin2018bert} and XLNet \cite{yang2019xlnet}, exploit this \emph{learnable lookup} to capture relations between words (or tokens). The attention layers, when inspected, can be seen to map real values to parts of the human-understandable input space (e.g., sentences). Exploration of the potential of the attention mechanism is becoming a lively research area on its own \cite{clark2019does,lin2019open}. We believe similar ideas can be investigated in the context of propositional (tabular) data
(where every row represents an individual data instance) that remain one of the most widely used data formats in academia and industry.

In this work, we propose the concept of \emph{Self-Attention Networks} (SANs) and explore, whether the representations they learn can be used for feature importance assessment, comparable to feature importance estimates returned by the established feature ranking approaches ReliefF \cite{robnik2003theoretical}, Mutual Information and Genie3 \cite{genie3,peng2005feature}.
The main contributions of this work are:

\begin{enumerate}
\item The Self-Attention Network (SAN) architecture, a novel neural network architecture capable of assessing feature importance directly, along with three ways of obtaining such importance estimates.
\item Extensive empirical evaluation of \textsc{SAN}s against ReliefF, Mutual Information and Genie3 feature ranking approaches, demonstrating comparable performance to the SAN architecture. The most important features according to SAN are shown to be in agreement with the ones detected by the established approaches used in the comparison.
\item Direct comparison of feature importance estimates, highlighting similarities between the considered algorithms' outputs. 
\item A theoretical study of SAN's properties, considering its space and time complexity. 
\end{enumerate}
In the remainder of this paper we first discuss the related work, followed by the formulation and empirical evaluation of SANs.

\section{Related work}
\label{sec:related}
In this section we present selected feature ranking approaches and briefly survey the approaches to learning from propositional data, followed by a presentation of neural attention mechanisms.

\subsection{Feature ranking}
\label{sec:fr}
\emph{Feature importance estimation} refers to the process of 
discovering parts of the input feature space, relevant for a given predictive modeling problem,
i.e. identifying the (most) important features to be used in model construction.

The simplest task of estimating feature importance is partitioning the features into groups of irrelevant and relevant ones, which is equivalent to assigning
every feature either importance $1$ (feature is relevant) or $0$ (feature is irrelevant). This task is referred to as \emph{feature selection}.
Here, one can only partially order the features.
In a more general case, when every feature is assigned an arbitrary importance score, one can sort the features with respect to these scores
and obtain \emph{feature ranking}. In this case, the features can be totally ordered.
Given a feature ranking, one can partition the features into relevant and irrelevant ones by thresholding, i.e., proclaiming
relevant the features whose importance is higher than some threshold.

The main motivation for performing feature ranking (or selection) is that an appropriately chosen subset of features may reduce the computational complexity of learning, while potentially increasing the learner's performance. Moreover, the learned models that use a smaller number of features are easier to explain.

In this work we consider a feature space $F$ with a corresponding class label set $C$.
We focus on feature ranking algorithms, but use thresholding (for many different values of the threshold) at the ranking-evaluation stage,
i.e., we measure the quality of the ranking as the predictive performance of the models that are built from sets of top-ranked features.

In this work we consider two types of feature ranking algorithms:
\begin{description}
\item[\textbf{Wrappers.}] A feature ranking algorithm of this family comes together with a supervised learner. Once trained on the training data, the learner is, apart from prediction, also capable of 
estimating how relevant is each of the features for the task at hand. An important member of this family is Genie3 feature ranking, computed directly from
Random Forest \cite{Breiman01a:jrnl}.
\item[\textbf{Filters.}] Algorithms such as the Relief family or Mutual Information do not need a predictive model to obtain feature rankings.
Since they are model-agnostic, the corresponding feature rankings are typically computed very efficiently, e.g., Mutual Information.
However, the computational efficiency often comes at the cost of myopia, i.e., they ignore possible feature interactions.
\end{description}
Recent advances in deep learning mostly address model-based ranking, where methodology, such as e.g., SHAP is used for \emph{post-hoc} analysis of a trained model. Note that such a methodology can also be model-agnostic, yet \emph{it needs} a model to compute feature importance.

\subsection{Propositional learning and neural networks}
In this work we focus on learning from 
(propositional) data tables, where columns represent features (attributes) and rows correspond to individual data instances.
Despite  its  simplicity,  training  neural network architectures on this type of data might be non-trivial in the cases of
small data sets, noisy feature spaces, spurious correlations, etc.
Further, recurrent or convolutional neural architectures are not that useful for tabular data as there is frequently no spatial or temporal information present in the data. Methods, such as Extreme Gradient Boosting \cite{chen2016xgboost} and similar, e.g., tree-based ensemble algorithms often dominate competitive shared tasks on such 
input spaces. 

Neural network approaches recently used in modeling propositional data are discussed next. The approach by Sakakibara \cite{sakakibara2005learning} addresses learning of context-free grammars based on tabular (propositional) representations. Further, multilayer perceptrons have been widely used, already in the previous century \cite{kemsley1992survey}. The recently introduced Regularization Learning Networks (RLNs) represent one of the most recent attempts at training deep neural networks on propositional data sets \cite{regnet}; the paper shows that correctly regularized neural networks perform on par with e.g., gradient boosting machines, demonstrating competitive performance on multiple data sets. Further, the authors also explored the information content of the feature space, showing that RLNs detect highly informative features. The attention-based propositional neural networks were also considered in the recently introduced TabNet \cite{arik2019tabnet}.
Finally, ensemble-based learners were also successfully applied to propositional data, including time series modeling \cite{qiu2014ensemble}. 

\subsection{Attention-based neural network architectures}
The area of deep learning has witnessed many novel ideas in recent years. The notion of \emph{attention} has emerged recently for language model learning \cite{devlin2018bert,vaswani2017attention}, as well as geometric deep learning \cite{wu2019comprehensive}. In short, the attention mechanism enables a neural architecture to pinpoint parts of the feature space that are especially relevant, and filter out the remainder.  The attention mechanism effectively sparsifies the input space, emphasizing only the elements relevant for the task at hand (e.g., language understanding). This way, the attention mechanism is commonly used to e.g., better learn the relations between words. Even though this mechanism has been widely adopted to text-based and graph-based tasks, less attention has been devoted to its application to propositional learning from tabular data.

\section{Self-Attention Networks}
\begin{figure*}[!t]
    \centering
    \includegraphics[width = 0.60\linewidth]{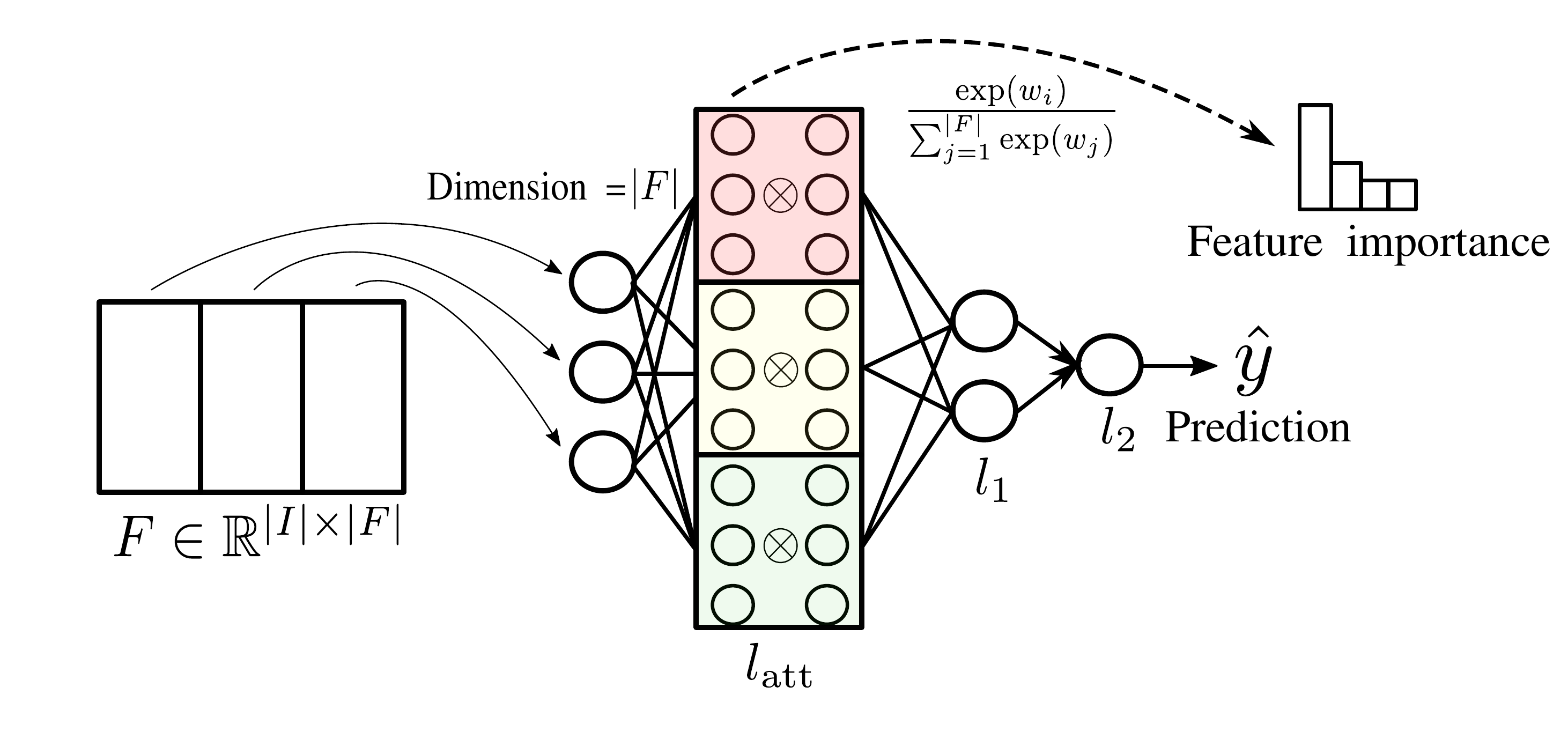}
    \caption{Overview of Self-Attention Networks. $l_{\textrm{att}}$ corresponds to the attention layer, where the element-wise product with the input space is computed on each forward pass. $l_1$ and $l_2$ correspond to two dense layers, yielding a prediction $\hat y$. The $\frac{1}{k} \oplus$  box represents element-wise means across the attention heads ($k$). Note that, if extracted, the attention vectors from $l_{\textrm{att}}$ can be used to compute the \emph{feature importance} estimates.}
    \label{fig:scheme}
\end{figure*}

This section presents the proposed Self-Attention Network (\textsc{SAN}) approach, illustrated in Figure~\ref{fig:scheme}. We begin by a formal description of the architecture, followed by feature importance computation.

\subsection{Propositional Self-Attention Networks}

This section sources some of the ideas from the seminal works on the attention mechanism \cite{chorowski2015attention,vaswani2017attention}. We refer the interested reader to the aforementioned publications for a detailed description and explain here only the essential ideas implemented as part of \textsc{SAN}s in a propositional learning setting. The neural network architecture that implements the attention mechanism can be stated as:
\begin{equation*}
    l_{2} = \sigma(W_{2} \cdot (a (W_{|F|} \cdot \Omega (X)+\bm{b}_{l_{1}})\\) + \bm{b}_{l_{2}}).
\end{equation*}
The first neural network layer $\Omega$ is designed specifically to maintain the connection with the input features $F$. We define it as:
\begin{equation*}
\Omega (X) = \frac{1}{k}\bigoplus_k \bigg [ X \otimes \textrm{softmax}( W_{l_\textrm{att}}^kX + \bm{b}_{l_{\textrm{att}}}^k) \bigg ].  
\end{equation*}
\noindent Input vectors $X$ are first used as input to a softmax-activated layer containing the number of neurons equal to the number of features $|F|$, where the softmax function applied to the $j_i$-th element of a weight vector $v$ is defined as:
\begin{align*}
    \textrm{softmax}(\bm{v}_{j_i}) = \frac{\exp(\bm{v}_{j_i})}{\sum_{j=1}^{|F|}\exp(\bm{v}_{j})}\text{,}
\end{align*}
where $v \in \mathbb{R}^{|F|}$. 
Note that $k$ represents the number of \emph{attention heads}---distinct matrices representing \emph{relations} between the input features.
The $\otimes$ sign corresponds to the Hadamard product and the $\oplus$ refers to the Hadamard summation across individual heads. $\Omega$ thus represents the first layer of a SAN, its output is of dimensionality $|F|$. $a$ corresponds to the activation function SELU \cite{klambauer2017self}, defined as:
\begin{equation*}
    \text{SELU}(x) = \lambda
    \begin{cases}
    \mbox{$x$} & \mbox{if } x > 0\\
    \mbox{$\alpha (\exp(x) - 1)$} & \mbox{if } x \leq 0
    \end{cases}\text{.}
\end{equation*}
where $\lambda$ and $\alpha$ are hyperparameters. The proposed architecture enables \emph{self-attention} at the feature level, intuitively understood as follows. As $\Omega$ maintains a \emph{bijection} between the set of features $F$ and the set of weights in individual heads $W_{l_{att}}^k$, the weights in the $|F|\times |F|$ weight matrix can be understood as \emph{relations between features}. 
Finally, SANs are additionally regularized using Dropout \cite{srivastava2014dropout}\footnote{For readability purposes, we omit the definition of this well known regularization method.}.
In this work, we are interested exclusively in self-attention---the relation of a given feature \emph{with itself}. We posit that such self-relations correspond to features' importance. Once a SAN is trained, we next discuss the ways of computing feature importance.

\subsection{Computing feature importance with \textsc{SAN}s}
\label{sec:dnx}
Let us show how the considered architecture can be used to obtain real values which we argue represent feature importance. We explore the procedures for obtaining the final vector, as a mapping from the feature space to the space of non-negative real values, i.e. function $f : F \to \mathbb{R}_0^{+}$.

\begin{description}
\item[Instance-level aggregations (attention).] Let
$(\bm{x}_i, \bm{y}_i)$, $1 \leq i\leq n$, be the
instances. Let \textsc{SAN}($\bm{x}_{i}$) represent the attention output of the $i$-th instance, defined as
\begin{equation*}
\textsc{SAN}(\bm{x}_i) = \frac{1}{k}\bigoplus_k \bigg [ \textrm{softmax}( W_{l_\textrm{att}}^k \bm{x}_i + \bm{b}_{l_{\textrm{att}}}) \bigg ].
\end{equation*}
The first option for computing feature importance performs the following operation, where the outputs are \emph{averaged} on-instance basis.
\begin{equation*}
R_{I} = \frac{1}{n}\sum_{i=1}^{n}\textsc{SAN}(\bm{x}_i).
\end{equation*}

\item[Counting only correctly predicted instances (attentionPositive).] The second variant of the mechanism operates under the hypothesis, that only correctly predicted instances should be taken into account, as they offer extraction of representative attention vectors with less noise.
Such scenarios are suitable when the number of instances based on which features' importance are to be computed is low, and the classification performance is not optimal.
Let $\hat{\bm{y}}_i$ 
represent the final prediction of an architecture (not only attention vectors).
This version of the approach assesses feature importance $R_I^{c}$ ($c$ stands for clean) as follows:

\begin{equation*}
R_I^{c} = \frac{1}{n}\sum_{i=1}^{n} \textsc{SAN}(\bm{x}_i)\left[\hat{\bm{y}}_i = \bm{y}_i \right]\text{.} 
\end{equation*}
\item [Global attention layer (attentionGlobal).] The previous two approaches construct the global feature importance vector incrementally, by aggregating the attention vectors based on individual instances used for training. However, such schemes can depend on the aggregation scheme used. The proposed global attention approach is more natural, as it omits the aggregation: once trained, we simply activate the attention layer's weights by using softmax. This scenario assumes, that the weight vector itself contains the information on feature importance, and can be inspected \textbf{directly}\footnote{In Appendix~\ref{appendixA} we discuss another way how global attention can be obtained, yet do not consider it in this work.}. Global attention is defined as follows:
\begin{equation*}
 R_G = \frac{1}{k}\bigoplus_k \bigg [ \textrm{softmax}( diag( W_{l_\textrm{att}}^{k})) \bigg ]; W_{l_\textrm{att}}^{k} \in \mathbb{R}^{|F| \times |F|}.
\end{equation*}

\end{description}

\section{Theoretical considerations}
\label{sec:theory}
In this section we discuss the relevant theoretical aspects of the considered neural network architecture, starting with SAN's space and time complexity, and followed by an overview of the computational complexity of the considered methods.

\subsection{Space and time complexity of SANs}
We first discuss the space complexity with respect to the number of parameters, as we believe this determines the usefulness of SANs in practice. Assuming the number of features to be $|F|$, the most computationally expensive part of SANs is the computation of the attention vector. The attention, as formulated in this work, operates in the space of the same dimensionality as the input space; the number of parameters is $\mathcal{O}(|F|^{2})$. This complexity holds if a single attention layer is considered. As theoretically, there can be multiple such mappings to the input space in a single \emph{attention block}, the space complexity in such case rises to $\mathcal{O}(|F|^{2} \cdot k)$, where $k$ is the number of considered attention weight matrices (heads). In practice, however, even very high dimensional data sets with, e.g., more than $60{,}000$ features can be processed, which we prove in the empirical section. Note that $(6 \cdot 10^{4})^{2} = 3.6 \cdot 10^{9}$ ($\approx$ 11GB if stored as floating points), which is the scale at which language models such as RoBERTa \cite{liu2019roberta} and similar operate, thus the complexity does not prohibit the use on high dimensional data that of practical relevance. We believe, however, that sparsity at the input level could significantly reduce the complexity, which we leave for further exploration.

We finally discuss the time complexity of computing feature importance for a single instance, as well as for a set of instances.
In Section~\ref{sec:dnx} we introduced three different implementations of SANs considered in this work: two instance-based, where attention is computed for each instance and aggregated afterwards, as well as global, where after training, the attention vector is simply extracted and used to represent features' importances. The first situation, where predictions up to the end of the second layer need to be conducted (up to the activated attention vector), is of linear complexity with respect to the number of test instances, across which the final importances shall be aggregated. On the other hand, if the attention vector is activated directly after the training phase, the computational complexity reduces to the computation of the softmax, and is in fact $\mathcal{O}(1)$ w.r.t. the number of test instances. Note that such computation is fundamentally faster and scales to massive production environments seamlessly.

\subsection{Overview of computational complexity}
\label{sec:complexity}
To contextualize the computational complexity of SANs, we compare it with some of the established feature ranking algorithms that will offer the reader insights into possible benefits, as well as drawbacks of individual methods considered in this work.

In Tables \ref{tbl:learn} and \ref{tbl:pred}, $t$ denotes the number of trees in the Random Forest, 
while $k$ denotes the number of attention matrices (heads) used.
 
\begin{table}
    \centering
        \caption{Computational complexity of feature importance estimation: The learning stage.} 
    \begin{tabular}{lcc}
    \hline 
      Algorithm   &  time complexity & space complexity \\ \hline
        ReliefF & $\mathcal{O}(|F| \cdot |I|^{2})$ & $\mathcal{O}(|F|)$ \\
        Random Forest & $\mathcal{O}(t|F| |I| \log^2 |I|)$ & $\mathcal{O}(|I| + |F|)$\\
        Mutual Information & $\mathcal{O}(|F| \cdot |I|)$ & $\mathcal{O}(|I|)$\\
        SAN & $\mathcal{O}(|F|^{2} \cdot |I|)$ & $\mathcal{O}(k \cdot |F|^{2})$\\
            \hline
    \end{tabular}
    \label{tbl:learn}
\end{table}

\begin{table}
    \centering
        \caption{Computational complexity of feature importance estimation: The post-learning stage. NA (not applicable) denotes the cases
        where feature importance is obtained directly from the learning stage.}
    \begin{tabular}{lcc}
        \hline 
      Algorithm   &  time complexity & space complexity \\ \hline
        ReliefF & NA &  NA \\
        Random Forest & $\mathcal{O}(t|I|)$ & $\mathcal{O}(|I|)$ \\
        Mutual Information & NA & NA \\
        SAN & $\mathcal{O}(1)$ or $\mathcal{O}(|I|)$ & $\mathcal{O}(k \cdot |F|^{2})$ \\
            \hline
    \end{tabular}
    \label{tbl:pred}
\end{table}
\noindent In Table~\ref{tbl:learn}, we compare the computational complexity of the learning process: this refers to either building a predictive model (Random Forest, SAN)
or computing the feature importances from the training data (ReliefF, Mutual Information). 
Further, in Table~\ref{tbl:pred}, we discuss the complexity of obtaining feature importances after the learning stage. Note that filter methods, such as the ReliefF and Mutual Information
only require the first step of the computation, whereas SAN and Random Forest require an additional computational step that extracts feature importances based on the trained model.

\section{Experimantal setting}
\label{sec:exp}
In this section we discuss the empirical evaluation setting used in this work. The evaluation is split into two parts: measuring the feature ranking similarity and
feature ranking quality. 
When measuring feature ranking similarity, we compare pairs of the rankings via \textsc{FUJI} score.
When measuring feature ranking quality, we explore how a series of classifiers behave, when top $n$ features are considered for the task of classification.

\subsection{Pairwise comparisons with \textsc{FUJI}}
\label{sec:fuji}
The output of a feature ranking can be defined as a real-valued list, whose $j$-th element is the estimated feature importance of the $j$-th feature.
A typical approach for comparing such lists is to compute the Jaccard indices \cite{Jaccard1901} between the sets of top-ranked features.
However, this score takes into account feature importances only implicitly (via the order of the features) and is consequently unstable and often too pessimistic.
Thus, we use its fuzzy version \textsc{FUJI} (the Fuzzy Jaccard Index) \cite{fujiMars}.

Even though \textsc{FUJI} takes importances directly into account, it is still scale-free and thus appropriate for comparing the outputs of different feature ranking algorithms.
Given a pair of feature rankings, e.g., Mutual Information and Attention, \textsc{FUJI} is computed for different sizes of top-ranked feature sets that belong to the two rankings.
The obtained values are then plotted as a single curve, as shown in Figure~\ref{fig:attendiff}. Such a comparison is very informative, since one can observe how the similarity
of the two rankings changes when the number of features considered grows, but aggregation of a curve into a single scalar may sometimes be preferred.
We do so by computing the area under the \textsc{FUJI} curve. Either trapezoidal or Simpson's rule can be employed;  in this work we use the latter\footnote{This implicitly assumes the \textsc{FUJI} curve is a spline of quadratic polynomials. We average the normalized areas across all data sets to obtain a single scalar representing the relation between two methods.}. 

In the experiments, we obtain feature rankings for similarity comparison as follows.
The model-based ranking algorithms (SANs and Random Forests) were trained by first selecting optimal model based on ten fold stratified cross validation.
This step is skipped in the case of ReliefF and Mutual Information algorithms. In the second step, we use the whole data set to estimate feature importances
(either directly or using the chosen model).
The whole data set can be used since these feature rankings are only used in the similarity computation (and not for predicting the target values),
and should be used since the feature rankings are expected to be better when computed from more data.

\subsection{Classification performance evaluation}
In the second set of the experiments, we investigate how the top $n$ features assessed by a given method influence the performance of the Logistic Regression (LR) classifier,
for different values of $n$. We selected this learner for the following reasons. As evaluation of individual values of $n$ requires hundreds (or thousands) of models to be built, LR is fast enough to be used in such setting. Next, it is sensitive to completely non-informative features. Should a given feature importance assessment method prioritize completely irrelevant features, the classifier will not perform well. In terms of hyperparameters, we use the same parametrization of SANs across all data sets. The rationale for this decision is that neural networks can easily overfit a given data set. We decided to use the same set of hyperparameters to showcase the overall adequate performance of SANs. The $l_1$ dimension was set to $128$, the network was trained for $32$ epochs, with the batch size of $5$. The learning rate for the Adam optimizer was set to $0.001$, and the dropout, which we used for regularization, was set to $20\%$. We used stratified $10$ fold cross validation where a feature ranking is computed from the training set and evaluated on the testing set. A single attention head was used ($k = 1$). We report the average Logistic Regression performance for each $n$.

\subsection{Experimental data sets}
\label{sec:datasets}
We conducted the experiments on the following data sets. 
\begin{description}
\item \textbf{DLBCL} \cite{armstrong2002mll}. A series of translocations specify a distinct gene expression profile that distinguishes a unique leukemia. The data set consists of 7{,}070 features and 77 instances.
\item \textbf{Genes} \cite{weinstein2013cancer}. The cancer genome atlas pan-cancer analysis project. The data set consists of 20{,}531 features and 801 instances. Features correspond to gene expression vectors.
\item \textbf{p53} \cite{danziger2009predicting}. Predicting Positive p53 Cancer Rescue Regions Using Most Informative Positive (MIP) Active Learning. The data set consists of 5{,}409 features and 16{,}772 instances. 
\item \textbf{Chess} \cite{shapiro1984role} (King-Rook vs. King-Pawn). The data set describes various endgames, and consists of 3{,}196 instances and 36 features.
\item \textbf{pd-speech} \cite{sakar2013collection}. Collection and Analysis of a Parkinson Speech data set with Multiple Types of Sound Recordings. The data set consists of 26 features and 1{,}040 instances.
\item \textbf{aps-failure} \cite{costa2016ida}. IDA 2016 Industrial Challenge: Using Machine Learning for Predicting Failures. The data set consists of 171 features and 60{,}000 instances.
\item \textbf{biodeg} \cite{dvzeroski1999experiments}. Biodegradability of commercial compounds. the data set consists of 62 features and 328 instances. 
\item \textbf{optdigits} \cite{alpaydin1998cascading}. Optical Recognition of Handwritten Digits. The data set consists of 64 features and 5{,}620 instances.
\item \textbf{madelon} \cite{guyon2005result}. This is a synthetic data set published at NIPS 2003, aimed to test feature selection capabilities of various algorithms. The data set consists of 500 features and 4{,}400 instances.
\end{description}
The selected data sets span across multiple topics, and were selected to provide insights into the quality of feature importance assessment methods. Finally, we also evaluated the difference in the attention across relevant and irrelevant features based on the algorithm proposed by Guyon et al. \cite{guyon2003design}\footnote{Implemented in \url{https://scikit-learn.org/stable/modules/generated/sklearn.datasets.make_classification.html}}. For the purpose of this synthetic experiment, we generated a binary classification problem data set comprised of 100 features, where only 50 were deemed relevant, with 1,000 samples of the data used for training SANs. 
The experiment was conducted via three repetitions of three-fold cross validation.
We aggregated the attention values separately for the positive, as well as the negative class if the classification accuracy for a given fold was more than 50\%. The attentions were aggregated only if a given prediction was correct---this step was considered to reduce the effect of the neural network's performance on the attention vectors' properties, as we only attempted to extract the attention relevant for the discrimination between the classes.

\section{Results of experiments}
\label{sec:results}
In this section, we present the results of the empirical evaluation. For improved readability, the results are presented graphically.

\subsection{Differences in attention vectors}
\begin{figure}[t]
    \centering
    \includegraphics[width = .5\linewidth]{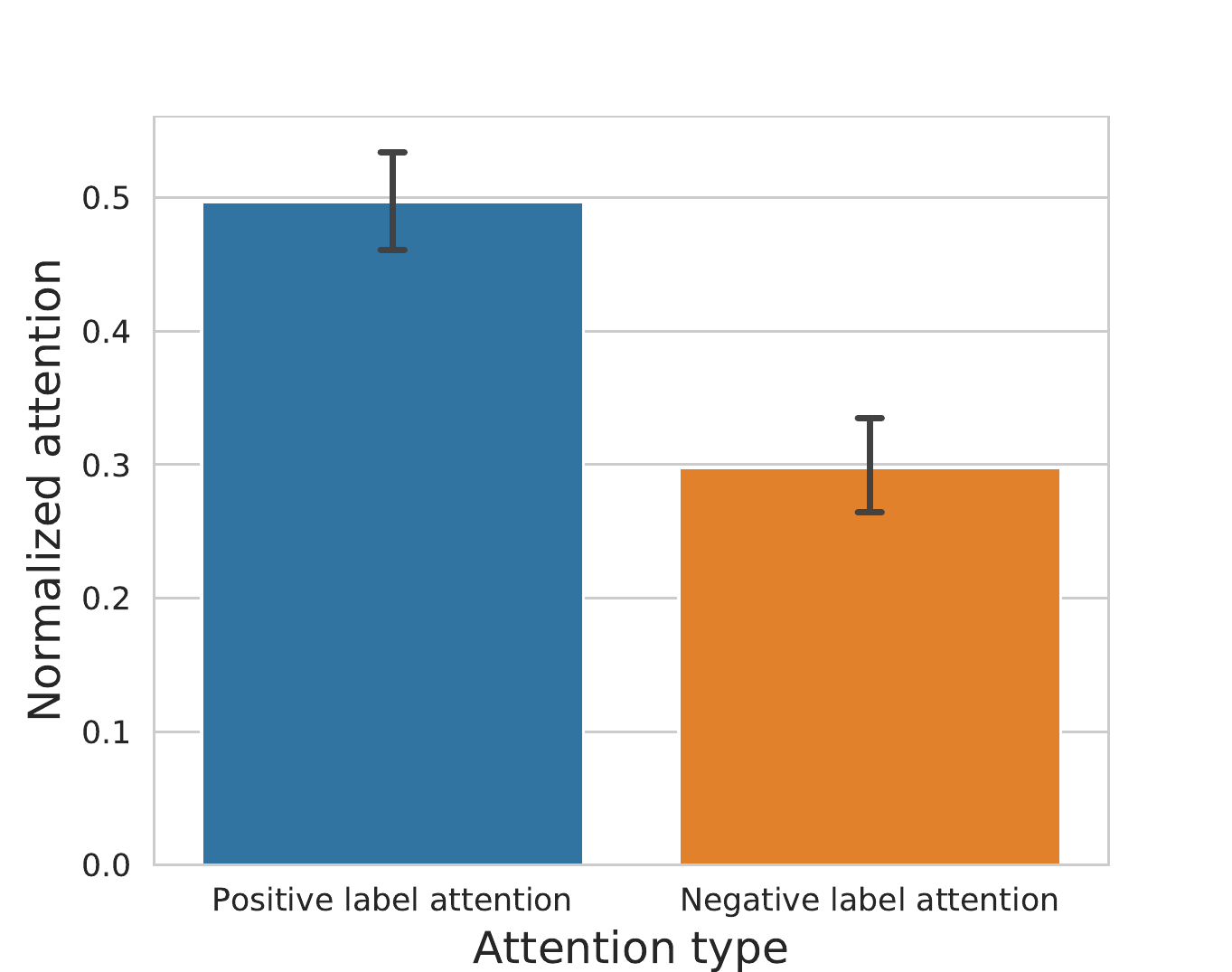}
    \caption{Differences in attention, aggregated w.r.t. correctly predicted positive and negative examples.}
    \label{fig:attendiff}
\end{figure}

We first visualize the distributions of attention with respect to both positive and negative instances in Figure~\ref{fig:attendiff}.

It can be observed that the attention is on average observably higher when positive instances are considered. As the neural network is trained to recognize such examples, such outcome indicates the attention mechanism potentially detects the relevant signal. Note that Figure~\ref{fig:attendiff}  shows attention, aggregated only on the features that were assessed as important (half of the features).

\begin{figure*}[h]
\centering
\resizebox{.9\textwidth}{!}{
\begin{tabular}{ccc}
{\includegraphics[width=0.99\linewidth]{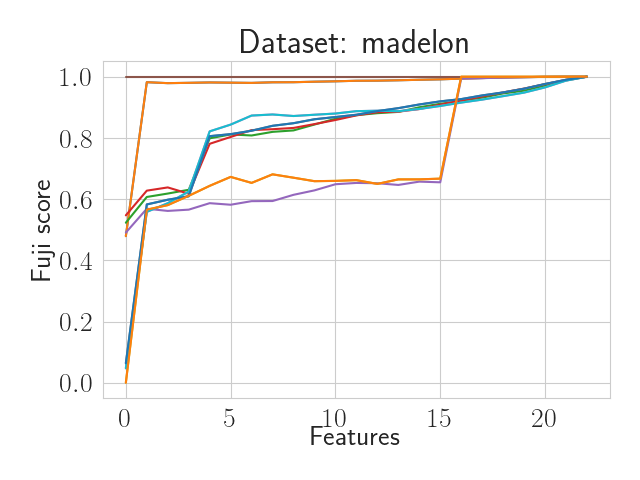}} &
{\includegraphics[width=0.99\linewidth]{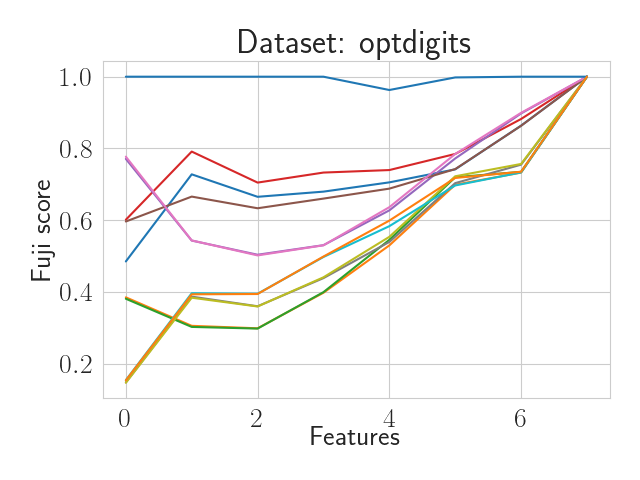}} &
{\includegraphics[width=0.99\linewidth]{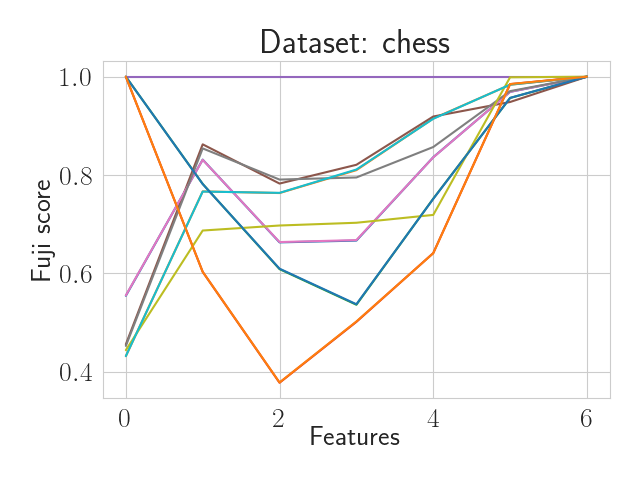}} \\
{\includegraphics[width=0.99\linewidth]{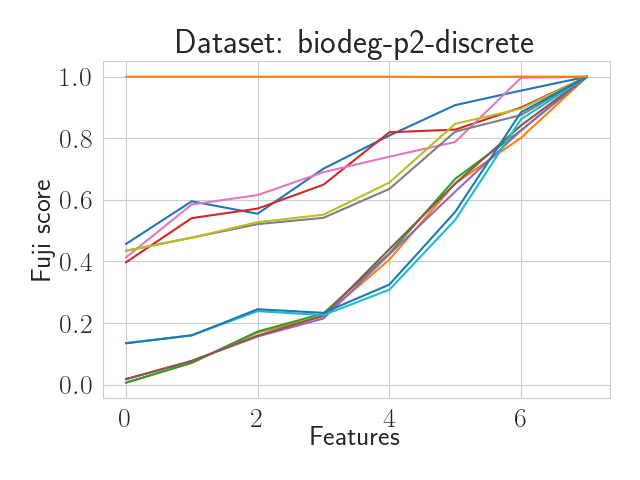}} &
{\includegraphics[width=0.99\linewidth]{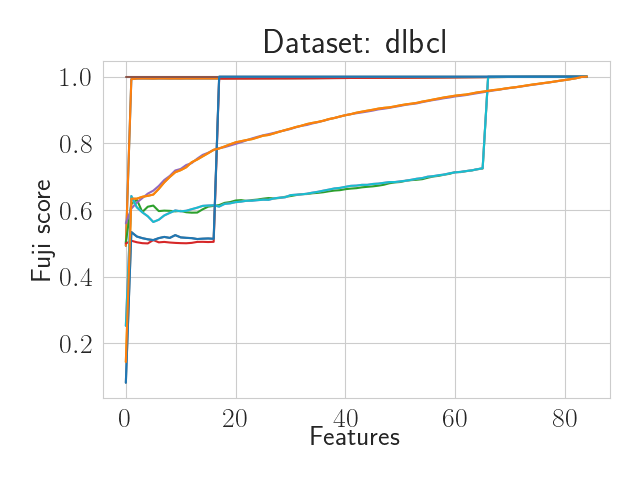}} & {\includegraphics[width=0.99\linewidth]{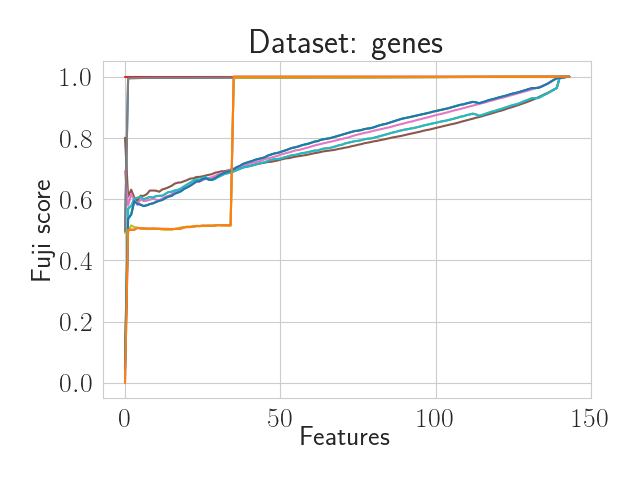}} \\  
{\includegraphics[width=0.99\linewidth]{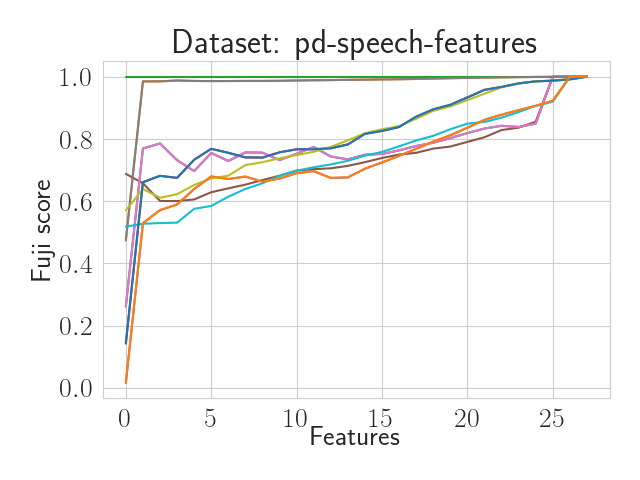}} &
{\includegraphics[width=0.99\linewidth]{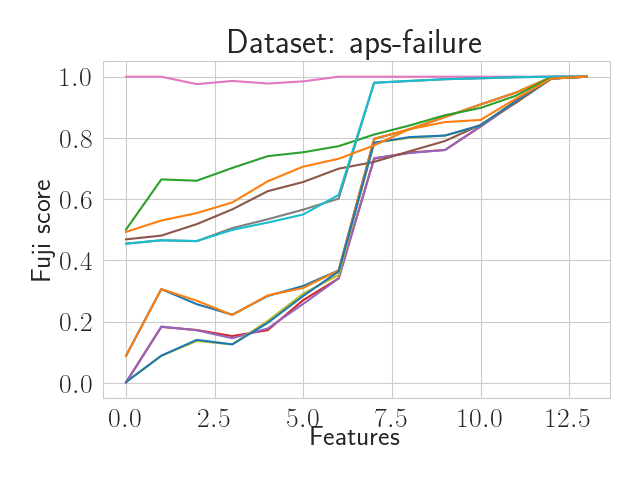}} & {\includegraphics[width=0.99\linewidth]{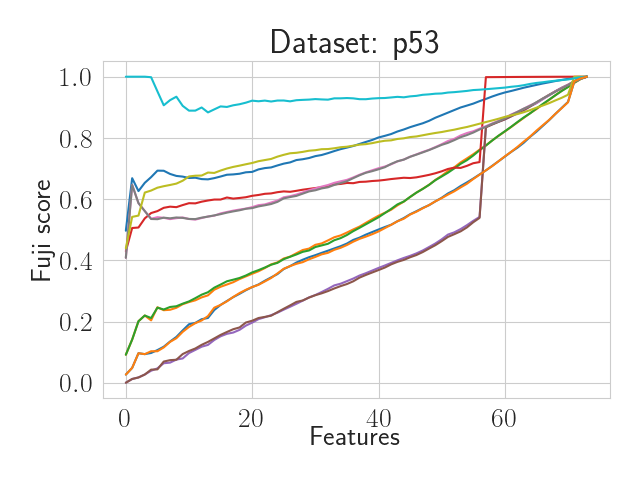}}  \\
\end{tabular}
}
\begin{tabular}{c}
{\includegraphics[width=0.89\linewidth]{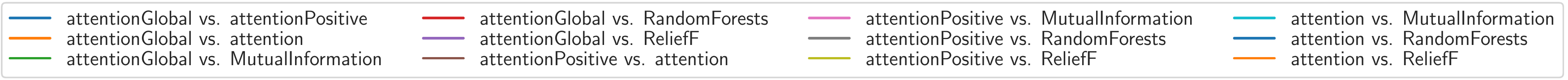}}
\end{tabular}
\caption{Results of \textsc{FUJI}-based ranking comparison. Higher values imply the pair of algorithms is more similar when a certain number of features (x-axis) is considered.}
\label{fig:lfr}
\end{figure*}
\subsection{Importance similarities}
\label{sec:similarities}

We next discuss the similarities between the considered feature importance approaches. The results are presented as \textsc{FUJI} curves, where we omit the space of all possible (pairwise) comparisons to the ones, that are compared with the proposed \textsc{SAN}s. The results are shown in Figure~\ref{fig:lfr}. 

\begin{figure}[h]
\centering
\includegraphics[width=.8\linewidth]{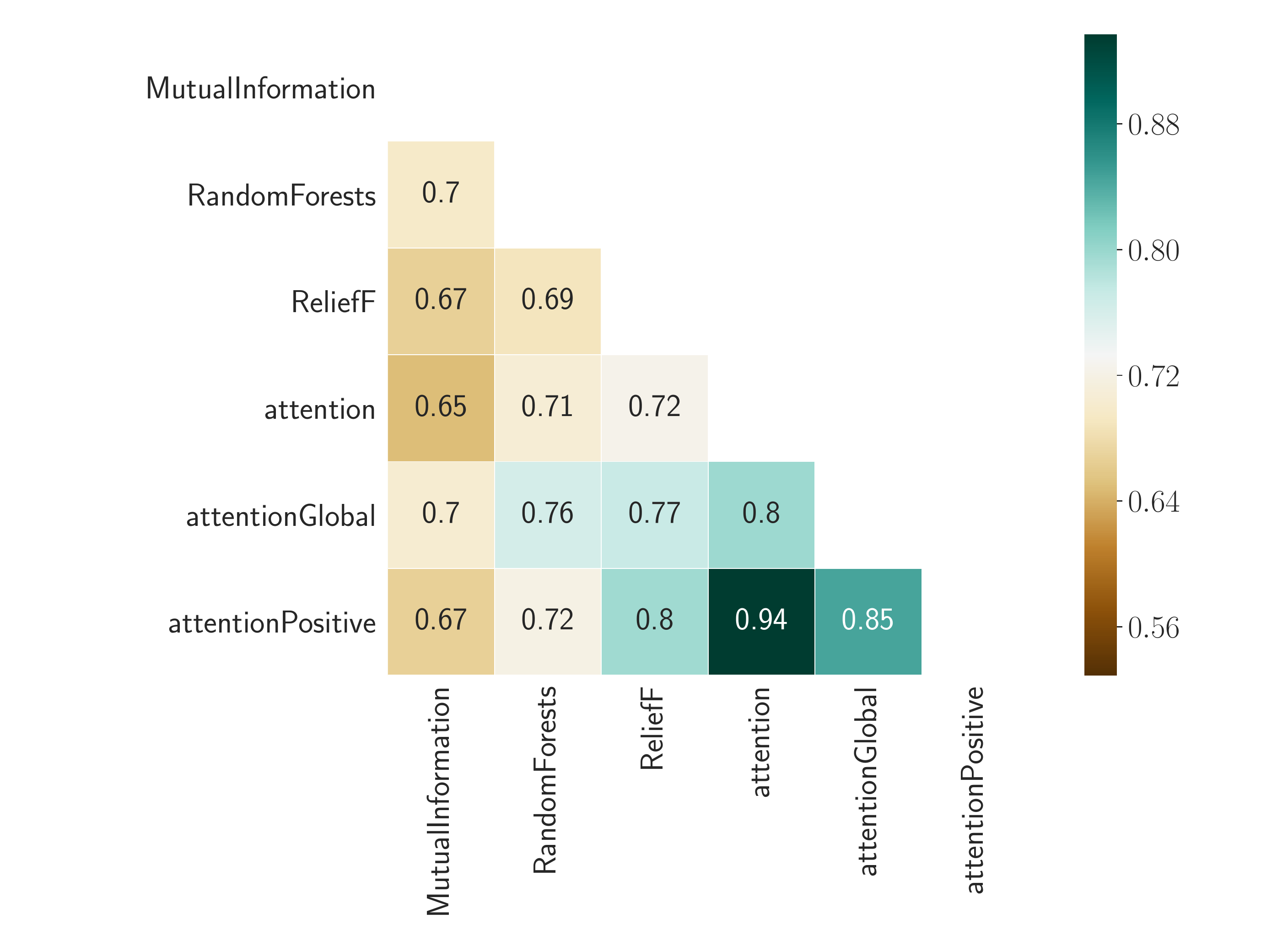}
\caption{Similarities of considered feature importance assessment approaches. The attention-based importances are the most similar to each other, as well as to the ReliefF algorithm.}
\label{fig:integrals}
\end{figure}

The considered visualizations depict at least two general patterns. First, the attentionClean and attention-based importances are indeed very similar. We remind the reader that the difference between the two is that attentionClean only takes into account the correctly classified instances. Next, we can observe that ReliefF and Random Forest-based importances are different. Further, the global attention appears the most similar to Random Forest importances. Overall areas below \textsc{FUJI} curves are shown in Figure~\ref{fig:integrals}.

\subsection{Classification performance}
\begin{figure*}[h]
\centering
\resizebox{.9\textwidth}{!}{
\begin{tabular}{ccc}
{\includegraphics[width=0.99\linewidth]{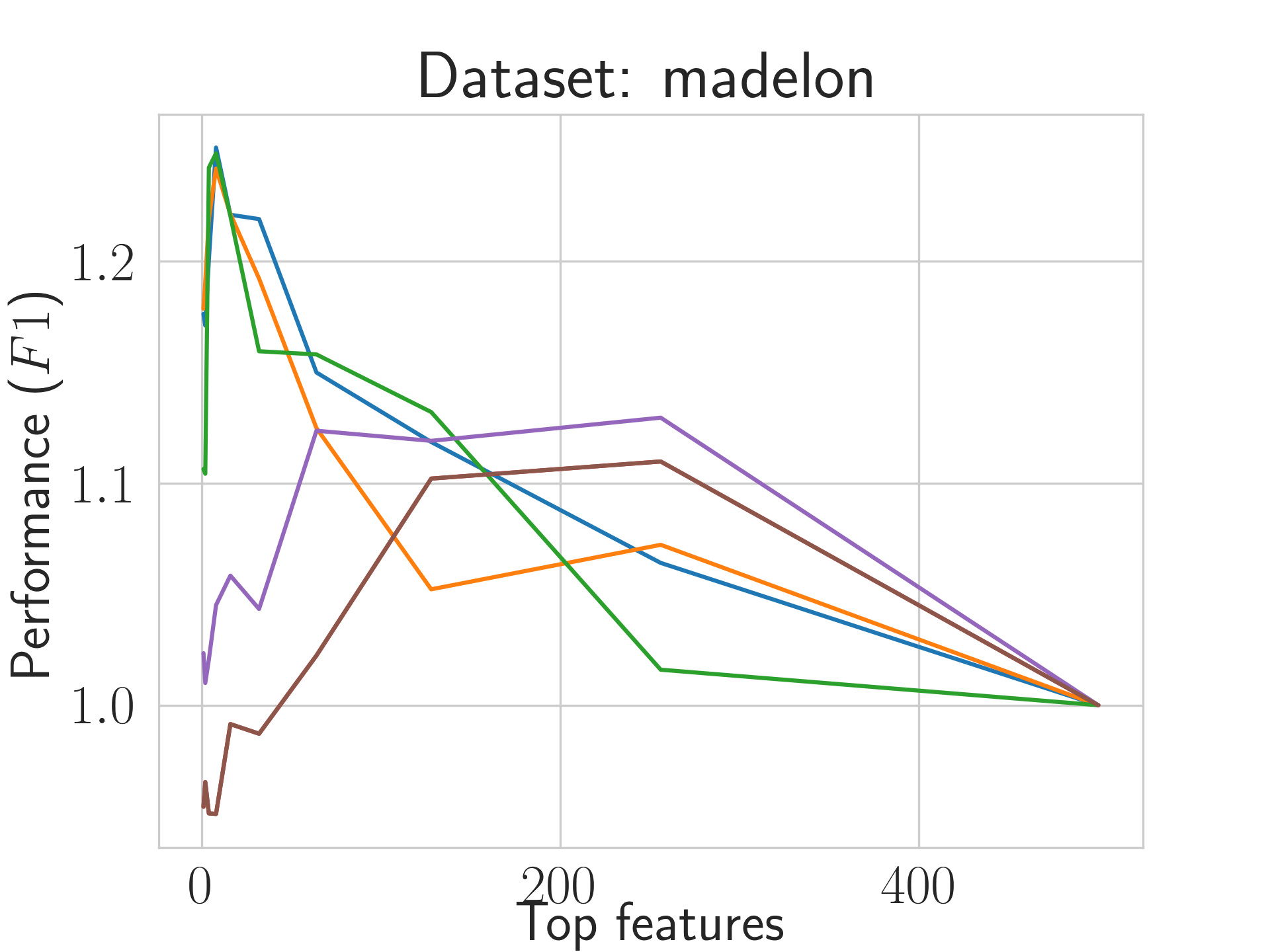}} &
{\includegraphics[width=0.99\linewidth]{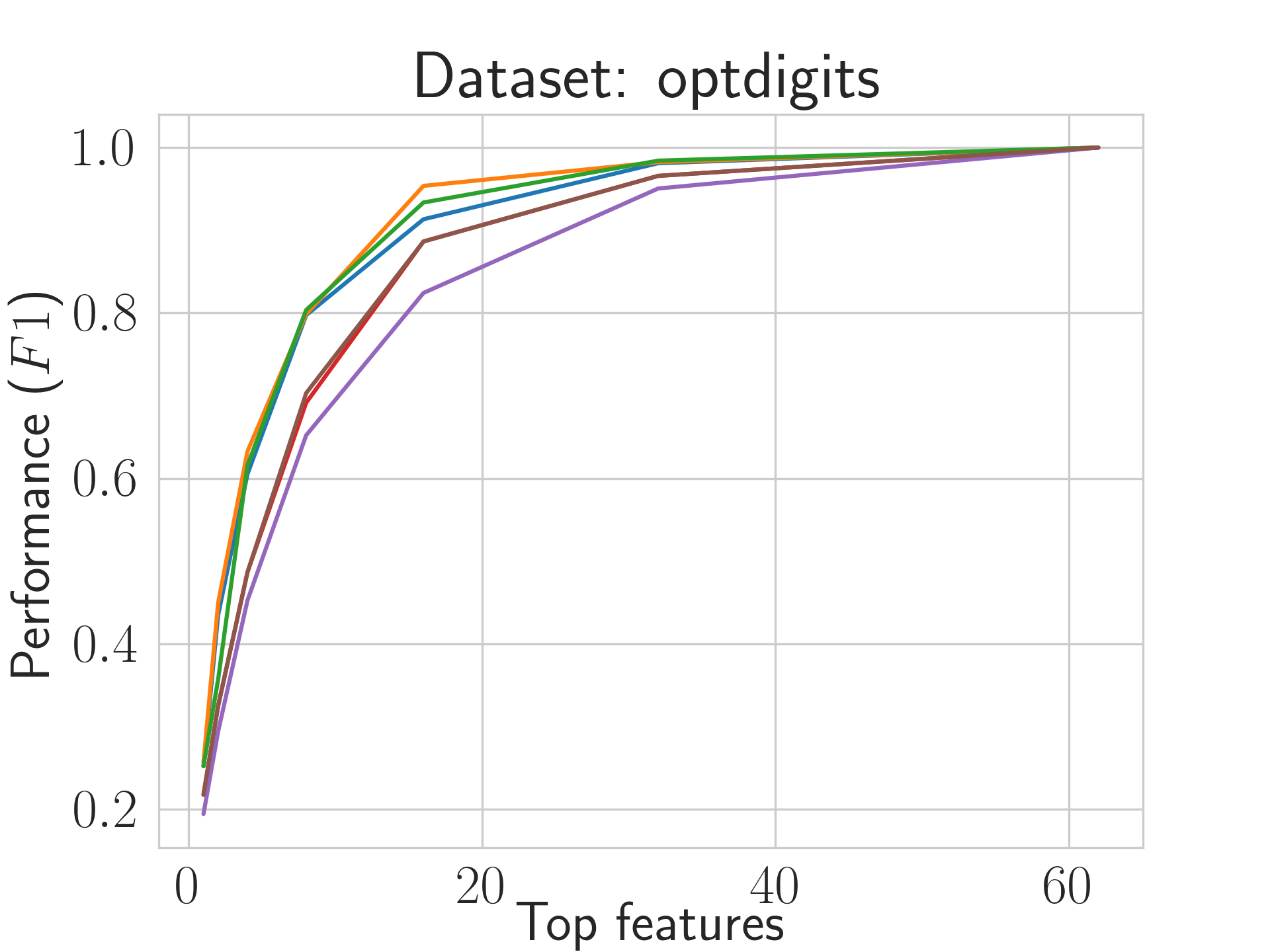}} &
{\includegraphics[width=0.99\linewidth]{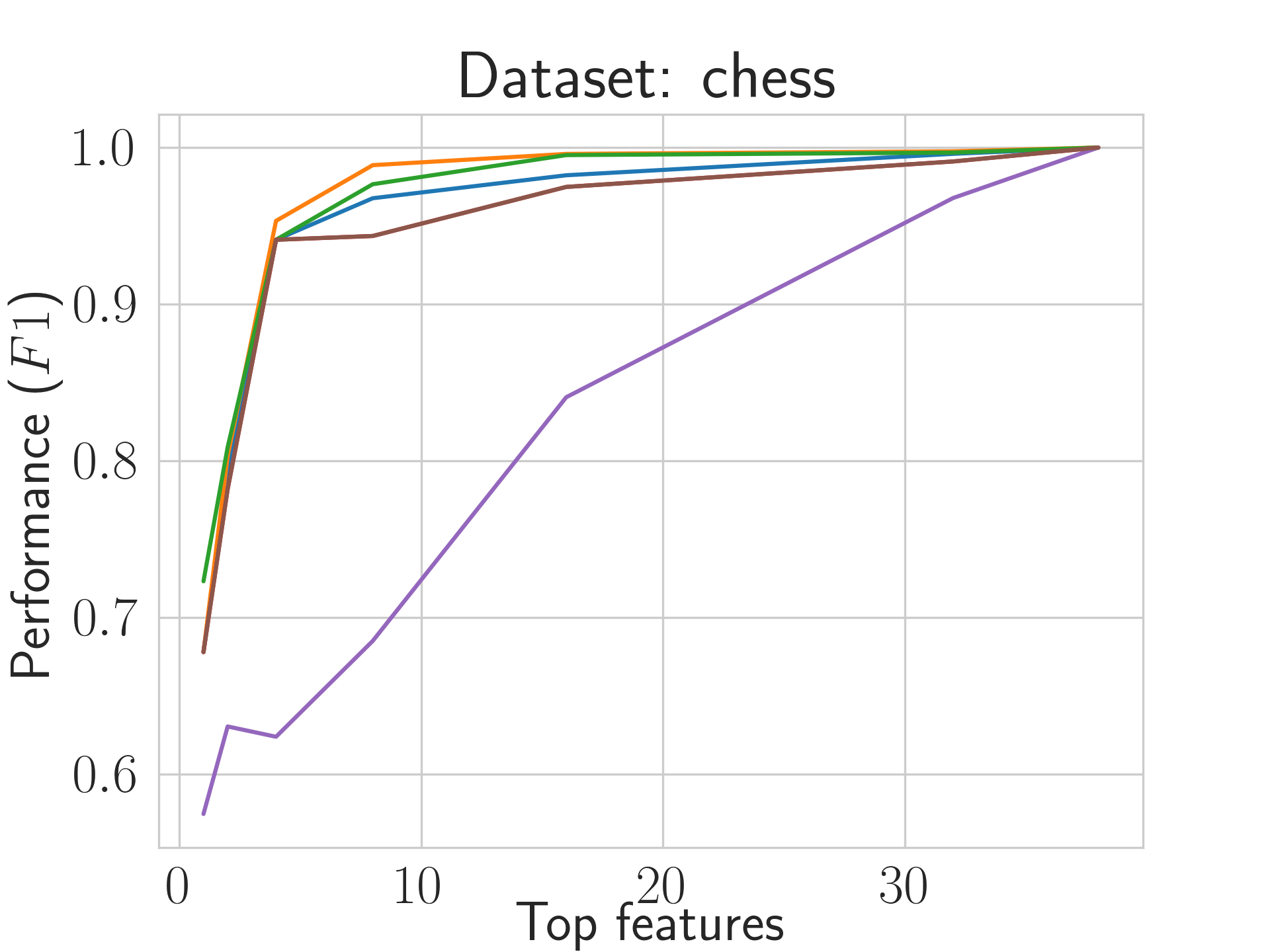}} \\
{\includegraphics[width=0.99\linewidth]{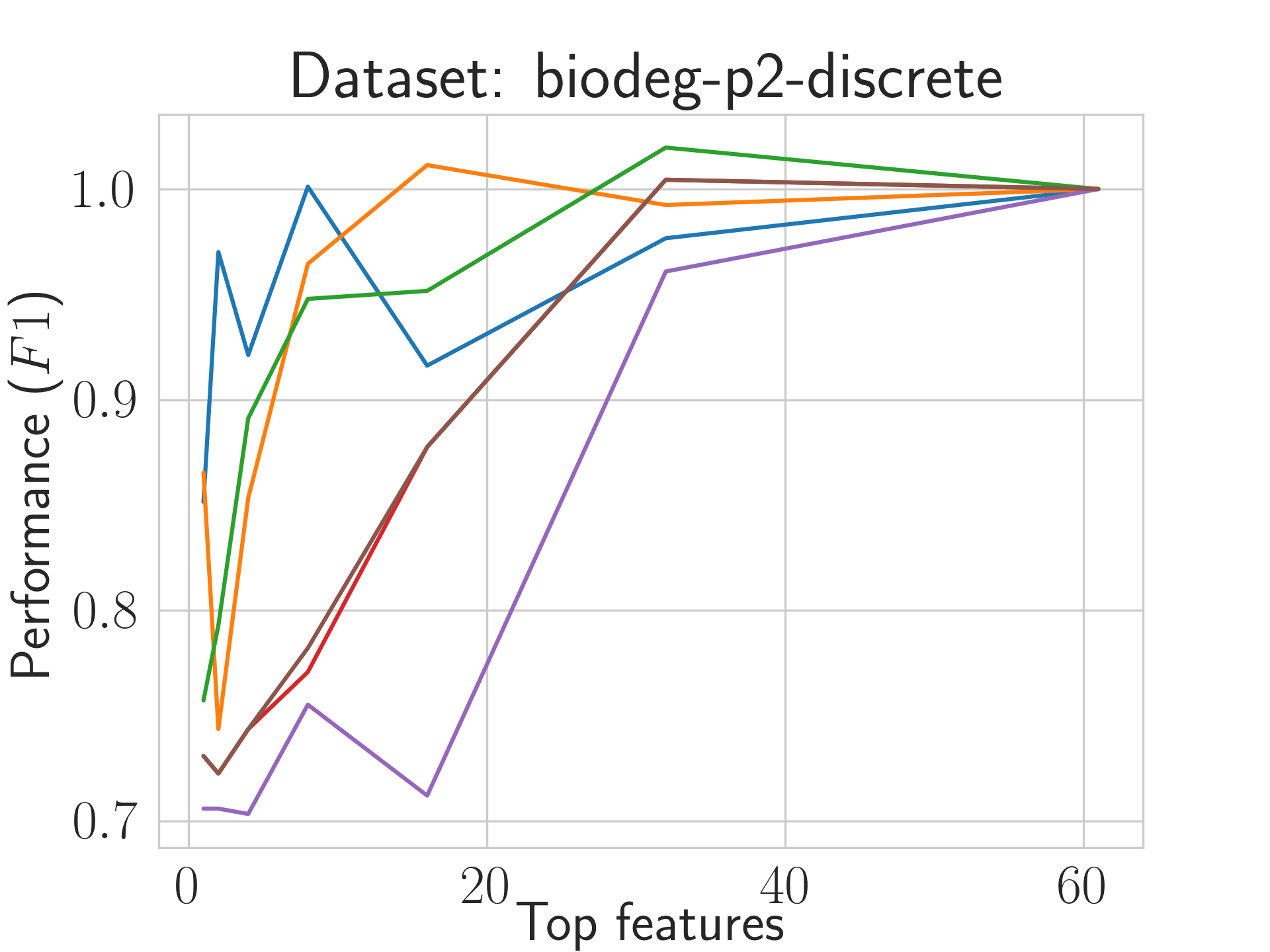}} &
{\includegraphics[width=0.99\linewidth]{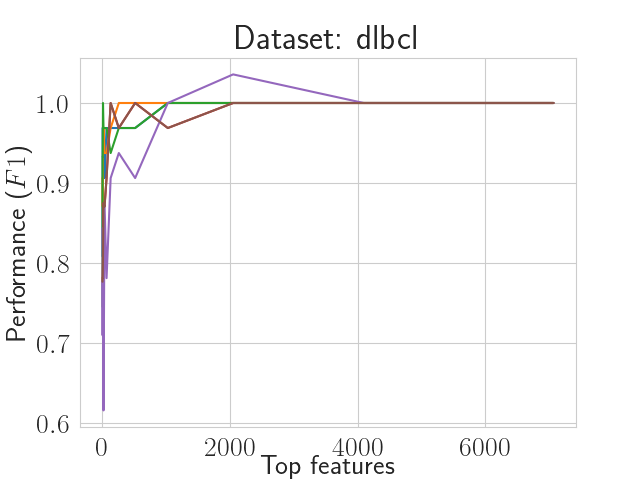}} & {\includegraphics[width=0.99\linewidth]{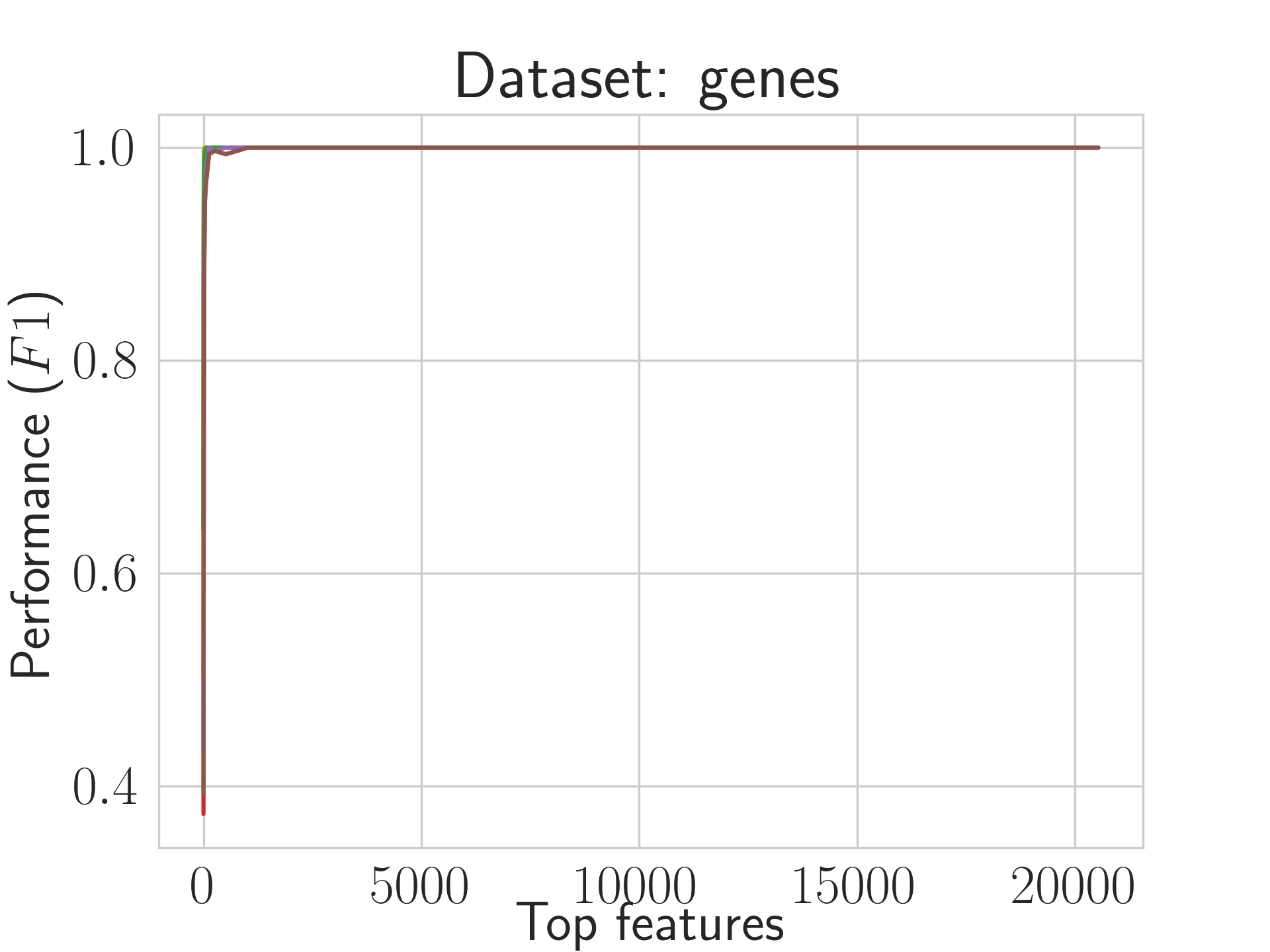}} \\
{\includegraphics[width=0.99\linewidth]{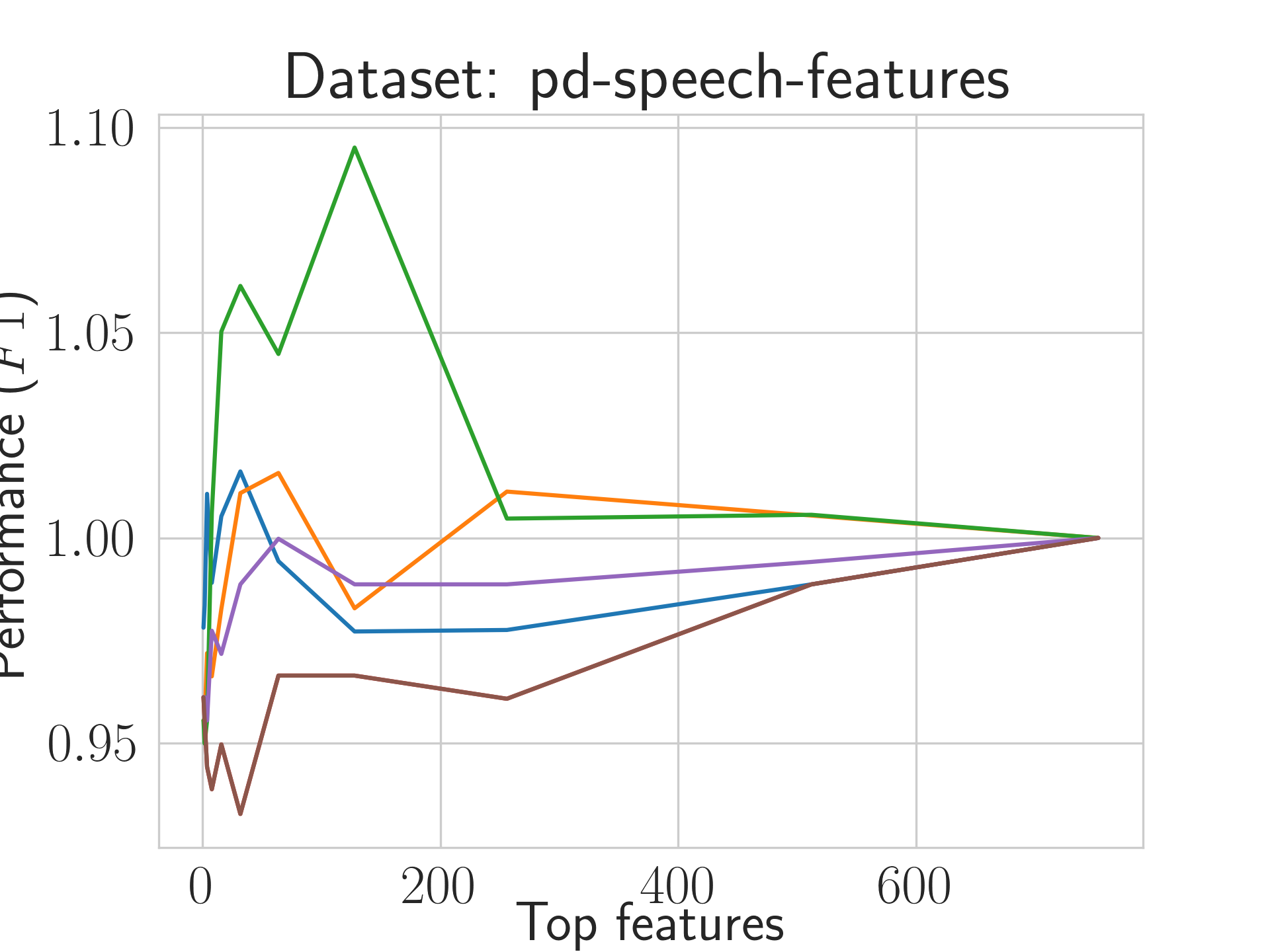}} &
{\includegraphics[width=0.99\linewidth]{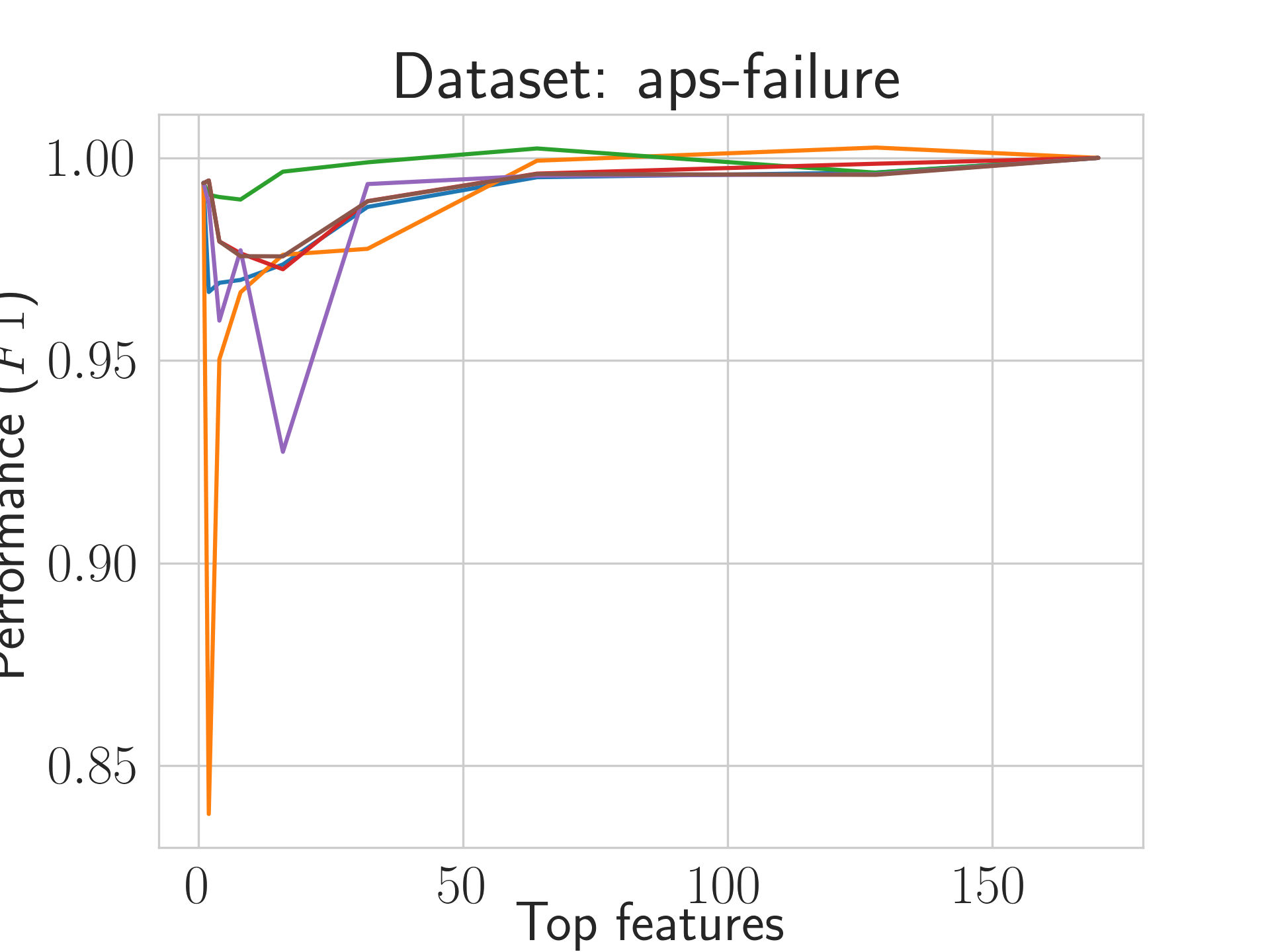}} &  {\includegraphics[width=0.99\linewidth]{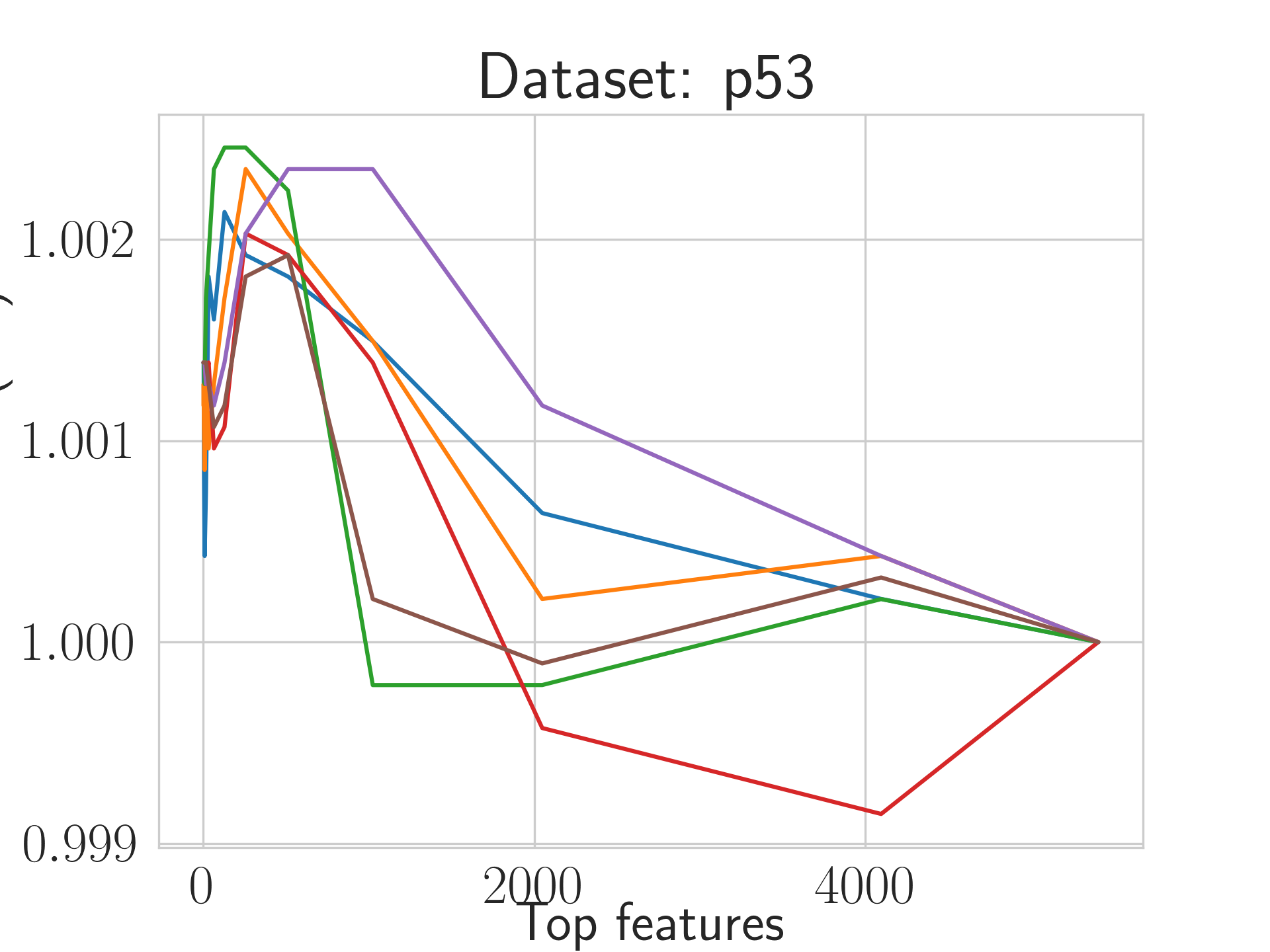}} \\
\end{tabular}
}
\begin{tabular}{c}
{\includegraphics[width=0.79\linewidth]{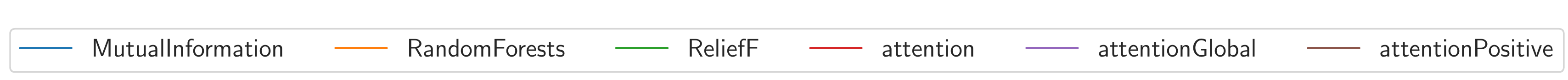}}
\end{tabular}
\caption{Feature ranking performance for different classifiers across different data sets. We report the \emph{relative} F1 score, i.e., performance with respect the setting where all features are used as inputs to the logistic regression classifier.}
\label{fig:clf}
\end{figure*}

We next discuss the classification performance results. Let us remind the reader that the results show how the logistic regression (C = 1, L2 normalization) classifier performs when parts of the feature space are pruned according to a given feature importance vector. The results are shown in Figure~\ref{fig:clf}.

The first observation---to our knowledge not known to the community---is that SANs indeed perform \emph{competitively}. Especially on data sets with a large number of features, such as the p53, genes and dlbcl (all biological ones), the attention mechanism detects larger subsets of features that can improve the performance of the logistic regression classifier, in e.g., dlbcl even outperforming all other approaches by a margin of more than 15\%. On smaller data sets, SANs perform competitively (e.g., madelon, optdigits, chess), or do not perform that well (biodeg-p2-discrete). Overall, worse performance of SANs coincides with smaller data sets, indicating that not enough data was present for SANs to distill the relevant parts of the feature space.
Further, it can be observed that ReliefF on data sets such as pd-speech and madelon outperforms all other approaches, indicating this state-of-the-art approach is suitable for low-resource situations.

\section{Discussion and conclusions}
\label{sec:discussion}


Operating on propositional data, \textsc{SAN}s have some inherent benefits, as well as drawbacks. The most apparent drawback is the architecture's space complexity, as with more attention heads, it could increase greatly for the data sets with a large number of features. In comparison, language models are maintaining a fixed sequence length during training, which is commonly below e.g., 1,024, and can thus afford multiple attention heads more easily.

Another key difference is the mapping to the input space. As with text, each input instance consists of potentially different tokens, attention matrices represent different parts of the vocabulary for each e.g., sentence. In contrast, the proposed \textsc{SAN}s exhibit a consistent mapping between the input space and the meaning of individual features, making possible the aggregation schemes considered in this work.

In terms of general performance, this paper proves empirically that SANs can emit attention vectors that offer similar feature rankings relevant for an external classifier (e.g., Logistic Regression). As the purpose of this work was not excessive grid search across possible architectures, which we leave for further work. We believe SANs's performance could be notably improved by improving the underlying neural network's performance, as well as by using other types and variations of the attention mechanism. Further, the proposed SANs shall be compared to the recently introduced 
TabNet \cite{arik2019tabnet}, an approach that also attempts to unveil the feature importance landscape via sequential attention mechanism.

An additional contribution of this work is the \textsc{FUJI} score based comparison, where we uncover similarities between \emph{all} considered approaches. The SANs are similar to the Genie3 and ReliefF, indicating the importance computation is possibly non-myopic (as MI), yet we leave validation of this claim for further work.

\label{sec:disc}

\section{Availability}
The code to reproduce the results is freely accessible for academic users at: \url{https://gitlab.com/skblaz/attentionrank}.
\ack We acknowledge the financial support from the Slovenian Research Agency through core research program P2-0103,
and project \emph{Semantic Data Mining for Linked Open Data} (financed under the ERC Complementary Scheme, N2-0078). The authors have received funding also from the European Union’s Horizon 2020 research and innovation program under grant agreement No 825153 (EMBEDDIA). The work of the first and the last author was funded by the Slovenian Research Agency through a young researcher grant.

\appendix
\section{Obtaining attention via feature-level activation}
\label{appendixA}
We defined global attention as:
\begin{equation*}
 R_g = \frac{1}{k}\bigoplus_k \bigg [ \textrm{softmax}( diag( W_{l_\textrm{att}}^{k})) \bigg ]; W_{l_\textrm{att}}^{k} \in \mathbb{R}^{|F| \times |F|}.
\end{equation*}
One could consider a similar scheme, with the difference that the softmax could be applied for each feature w.r.t. the remainder of the features, and simply extracted as the main diagonal of the obtained matrix. Let $\textrm{RWS}$ be defined as:
\begin{equation*}
\textrm{RWS} = \textrm{concatByRows} (\textrm{softmax} (W_{l_\textrm{att}}(i) ));\textrm{i is a row}.
\end{equation*}
Where $W_{l_\textrm{att}}(i)$ corresponds to the i-th row of the weight matrix.
Thus, the global attention can be computed as:
\begin{equation*}
 R_g = \frac{1}{k}\bigoplus_k \bigg [  diag(\textrm{RWS}( W_{l_\textrm{att}}^{k})) \bigg ].
\end{equation*}

\bibliography{ecai}

\end{document}